\documentclass[letterpaper,times]{IONconf-v2}
\usepackage[hidelinks]{hyperref}

\newcommand{\eg}{\emph{e.g.},}
\newcommand{\ie}{\emph{i.e.},}

\newcommand{\datasetname}{In-house}
\newcommand{\gnetss}{GNetSS}
\usepackage{threeparttable}
\usepackage{siunitx}

\articletype{Original paper}%

\journalname{NAVIGATION}
\journalvolume{}
\journalnumber{}

\title{GNetSS: Towards Learning GNSS Corrections\\ without Ground Truth}

\author[1,2]{JiaXu Chee}

\author[2]{Simon Denman}

\author[1]{Fahira Afzal Maken}

\author[2]{Qingqing Cheng}

\author[3]{Amir Khodabandeh}

\author[1]{Milad Ramezani}
\authormark{CHEE \textsc{et al}}

\address[1]{Technology, CSIRO, Queensland, Australia}

\address[2]{Faculty of Engineering, Queensland University of Technology, Queensland, Australia}

\address[3]{Department of Infrastructure Engineering, The University of Melbourne, Queensland, Australia}

\corres{JiaXu Chee\\ Technology, CSIRO, Queensland, Australia.\\ \email{jiaxu.chee@csiro.au}}

\begin{document}

\abstract[Abstract]{We present \emph{\gnetss}, a self-supervised GNSS position correction framework that improves the robustness of single point positioning (SPP) without requiring ground-truth locations for training. \gnetss{} combines a transformer-based network with a deterministic differentiable positioning solver. The network predicts satellite-wise pseudorange corrections by modelling inter-satellite geometry and temporal context, while the solver converts these corrected measurements into receiver position corrections under GNSS geometry and optional motion constraints. Unlike supervised GNSS correction methods that rely on ground-truth, \gnetss{} derives its training signal directly from GNSS measurement consistency.
Experimental results show that \gnetss{} improves robustness across open-sky, urban, and forested environments. The method is particularly effective at suppressing large-error outliers, reducing the maximum 3D positioning error by over 50$\%$ compared with SPP. These results demonstrate the potential of self-supervised pseudorange correction for scalable and robust GNSS positioning in real-world autonomy applications. Code and the in-house dataset will be released upon acceptance.}


\maketitle

\section{Introduction}
\label{sec:introduction}

Accurate global positioning is a fundamental requirement for autonomous robotic systems, particularly in applications such as autonomous driving~\citep{Bresson2017}, field robotics~\citep{robotics9040097}, and large-scale navigation~\citep{yin2024survey}. Global Navigation Satellite Systems (GNSS) provide a widely available solution for global localisation. However, their accuracy is often degraded in real-world environments due to factors such as multipath interference, non-line-of-sight (NLOS) reception, atmospheric disturbances, receiver noise, and poor satellite geometry. These challenges are especially pronounced in urban canyons, forested areas, and other cluttered environments, where reflected or blocked signals introduce biased pseudorange measurements that cannot be fully compensated for by conventional stand-alone GNSS positioning solutions~\citep{kaplan2017gnsserrors}.

Classical GNSS positioning methods (\eg~SPP~\citep{teunissen2017springer}) rely on   
physics-based correction models to account for satellite clock bias, atmospheric delays, the Earth's rotation, and other systematic effects \citep{kaplan2017gnsserrors}. While such model-based approaches are interpretable and effective under favourable assumptions, they remain vulnerable to environment-dependent errors such as multipath and NLOS reception. 

Learning-based GNSS correction methods have recently shown promise in modelling complex pseudorange errors that are difficult to describe analytically. Existing approaches have been used to classify line-of-sight (LOS)/NLOS signals, estimate pseudorange errors, predict measurement weights, or directly regress position corrections. However, most of these methods are supervised and require ground-truth labels obtained from real-time kinematic (RTK), simultaneous localisation and mapping (SLAM) or 3D city models. This dependency limits scalability, since collecting accurate reference trajectories across diverse environments, receivers, and satellite conditions is expensive and often impractical.

\begin{figure}[t]   
    \centering   
    \begin{subfigure}[b]{0.5\linewidth}     
    \centering     
    \includegraphics[width=\columnwidth]{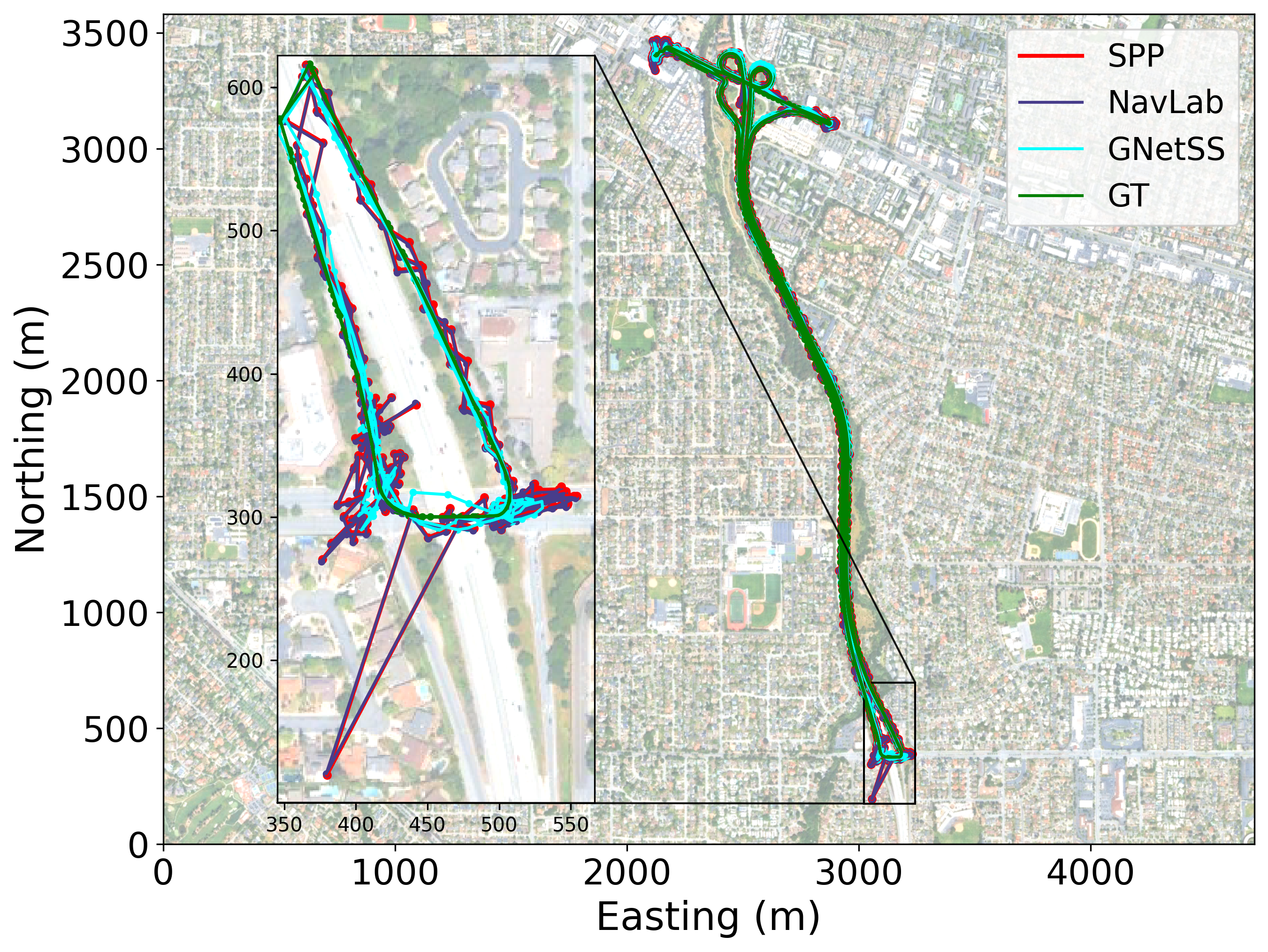}
    \end{subfigure}\par
    
    \begin{subfigure}[b]{0.5\linewidth}     
    \centering     
    \includegraphics[width=\columnwidth]{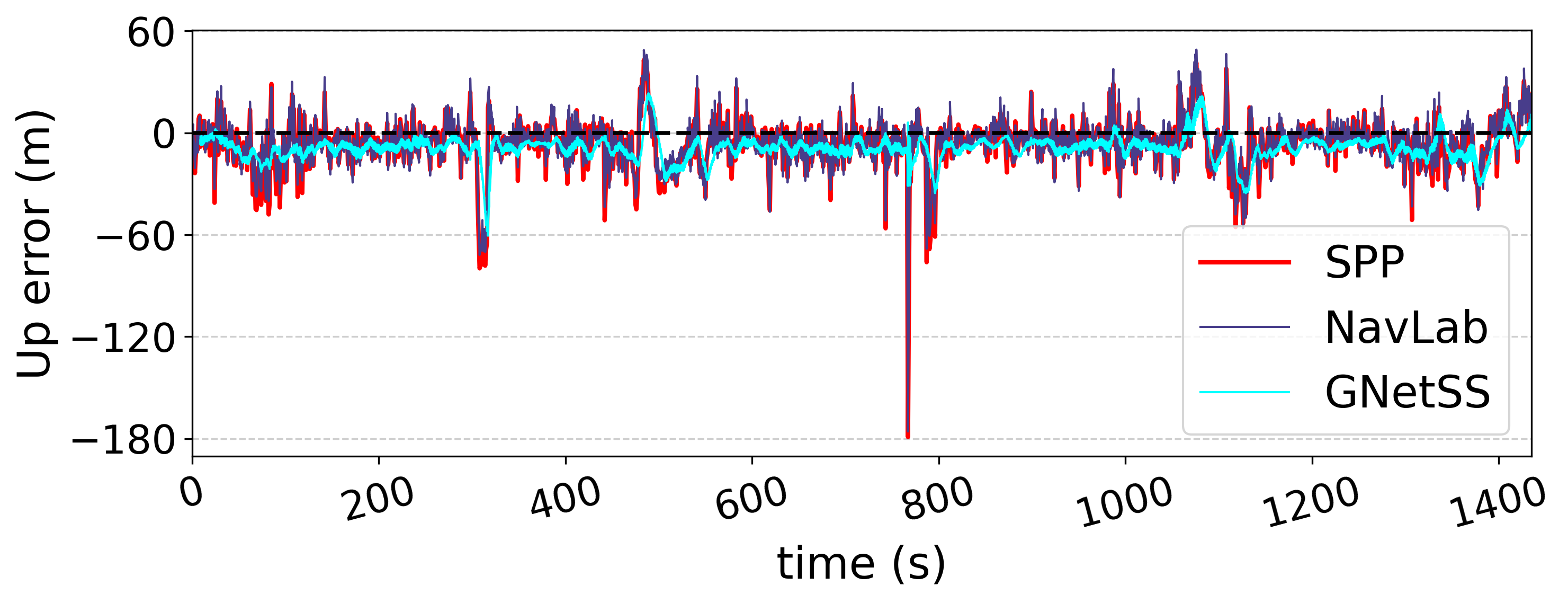}
    \end{subfigure}   
    
    \caption{Example positioning results in an urban area, comparing~\gnetss{} with the baselines: estimated horizontal trajectories in Easting-Northing (top) and corresponding vertical errors (bottom).}   
    \label{fig:hero} 
\end{figure}

In this article, we propose \textbf{\gnetss}, a \textbf{G}NSS \textbf{Net}work for \textbf{S}elf-\textbf{S}upervised position correction. \gnetss{} learns satellite-wise pseudorange corrections directly from GNSS-derived measurements without requiring ground-truth receiver positions. The proposed framework combines a transformer-based correction network with a deterministic differentiable positioning solver. The network models inter-satellite relationships at each epoch and temporal context across sequential observations, while the solver enforces the geometric structure of GNSS positioning by converting learned measurement corrections into receiver position corrections. Training is performed using a self-supervised objective, allowing the model to learn from GNSS measurement geometry and temporal consistency rather than external position labels.

The design of \gnetss{} follows a hybrid learning-and-model-based formulation that retains the strengths of GNSS geometric positioning while improving adaptability to environment-dependent errors. Rather than replacing the positioning solver with a direct position regressor, \gnetss{} learns corrections at the measurement level and passes them through the solver. This keeps the correction process tied to the underlying GNSS measurement geometry and makes the learned outputs physically interpretable. In addition, the self-supervised formulation allows the model to exploit raw GNSS sequences collected across diverse environments, while the dynamic solver encourages temporally consistent trajectory estimates.

Experimental results show that the proposed method is particularly effective at suppressing large-error outliers, as illustrated in Figure~\ref{fig:hero}, achieving the lowest or near-lowest maximum 2D and 3D positioning errors in most cases. 

The main contributions of this work are as follows:
\begin{itemize}
\item We introduce \gnetss, a self-supervised GNSS position correction framework that learns pseudorange corrections without ground-truth receiver positions.
\item We propose a transformer-based correction network coupled with a deterministic differentiable positioning solver, preserving GNSS measurement geometry while enabling end-to-end learning.
\item We formulate a self-supervised training objective that enables learning from raw GNSS sequences using measurement and trajectory consistency.
\item We validate the proposed method on public and in-house datasets, and demonstrate improved robustness to large positioning outliers in challenging environments.
\end{itemize}

\section{Related Works}
\label{sec:relatedworks}
This section reviews model-based and data-driven GNSS positioning methods relevant to this work. We focus on approaches using single-receiver code measurements.

\subsection{Model-Based Methods}
\label{sec:rw_mb}
Model-based methods formulate positioning as a state estimation problem using explicit measurement~\citep{misra2006global}, motion~\citep{4792993}, and noise models~\citep{weng2023localization}. These approaches encode satellite geometry and receiver dynamics in an interpretable way, but are sensitive to factors such as the validity of the assumed models.

Recursive filtering and smoothing methods have also been applied to raw GNSS measurements. \citet{weng2023localization} applied Extended Kalman Filtering (EKF)~\citep{grewalKalmanFilteringTheory2008}, Moving Horizon Estimation (MHE)~\citep{4792993}, and Rauch--Tung--Striebel (RTS)~\citep{rts1965} smoothing to raw GNSS measurements obtained from Android devices~\citep{fu2020android_raw_gnss}. Among these, EKF and MHE are suitable for real-time applications, whereas RTS smoothing requires a backward pass over the sequence and is thus best suited to post-processing. 

Recent works have adopted factor-graph frameworks for GNSS positioning~\citep{wen2021towards,suzuki2025open}, which optimise over historical states and measurements rather than relying solely on recursive filtering. This can improve temporal consistency, but generally comes at a higher computational cost. In practice, libraries such as GTSAM~\citep{dellaert2012factor} support fixed-lag smoothing, where optimisation is performed over a batch or sliding window of recent states, while older states are marginalised as new measurements arrive.



Overall, model-based GNSS methods provide interpretable and computationally efficient position estimates. However, their performance can deteriorate in the presence of complex, environment-dependent errors such as multipath propagation and NLOS reception. These challenges have motivated the development of data-driven approaches.

\subsection{Data-Driven Methods}
\label{sec:rw_dl}

Data-driven GNSS positioning methods have used machine learning, and most recently deep learning (DL), to learn complex and environment-dependent error patterns that model-based methods can struggle to capture. 

A common line of work focuses on classifying GNSS measurements according to their propagation condition using classifiers such as support vector machines (SVM)~\citep{Hsu2017GNSSApproach, Lyu2020AnEnvironments, Suzuki2021NlosLearning} and neural-networks (\eg~multi-layer perceptron (MLP))~\citep{Suzuki2021NlosLearning} for NLOS detection.
While these methods demonstrate that learning-based classifiers can identify degraded measurements, their dependence on auxiliary sensors
~\citep{Lyu2020AnEnvironments,Suzuki2021NlosLearning}  or detailed environmental models for label generation~\citep{Hsu2017GNSSApproach} makes large-scale deployment challenging.

Beyond signal classification, several works learn to estimate pseudorange errors directly. \citet{Sun2021ImprovingLearning} used a gradient boosting decision tree (GBDT) in a static environment to predict pseudorange errors, which were either used directly to correct measurements or combined with deterministic NLOS categorisation and exclusion. They found that excluding NLOS measurements degraded accuracy because the reduced number of visible satellites weakens satellite geometry. \citet{Sun2023ResilientAreas} later introduced a random-forest approach that predicts and corrects pseudorange errors from dynamic GNSS data.
\citet{Zhang2021LSTM} combined long short-term memory (LSTM)~\citep{hochreiter1997long} and fully connected neural networks to predict satellite visibility and pseudorange error; while \citet{Weng2024PrNet} used an MLP with RTS smoothing to refine the pseudorange measurements. Since RTS smoothing requires backward processing, the resulting pipeline is better suited for post-processed trajectory estimation than real-time positioning. \citet{Hu2025TDL} used an MLP to predict both pseudorange corrections and weights for weighted least squares (WLS), allowing the positioning solver to account for learned measurement reliability.

\begin{figure*}[t]
\centering
\begin{subfigure}[b]{0.49\linewidth}
    \centering
    \includegraphics[width=0.95\linewidth]{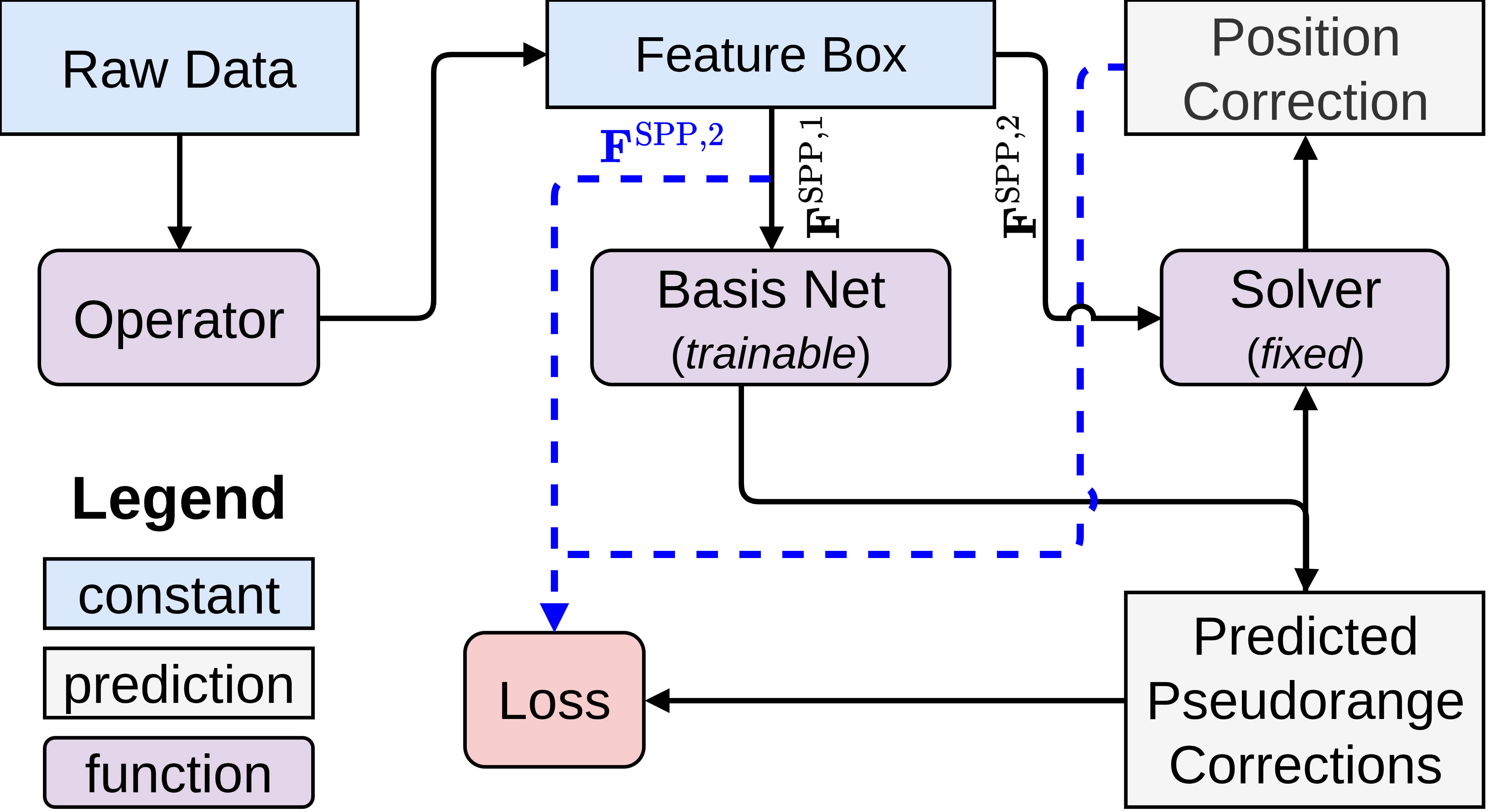}
    \caption{Training pipeline via self-supervised learning (SSL).}
    \label{subfig:training_pipeline_SSL} 
\end{subfigure}
\hfill
\begin{subfigure}[b]{0.49\linewidth}
    \centering
    \includegraphics[width=0.95\linewidth]{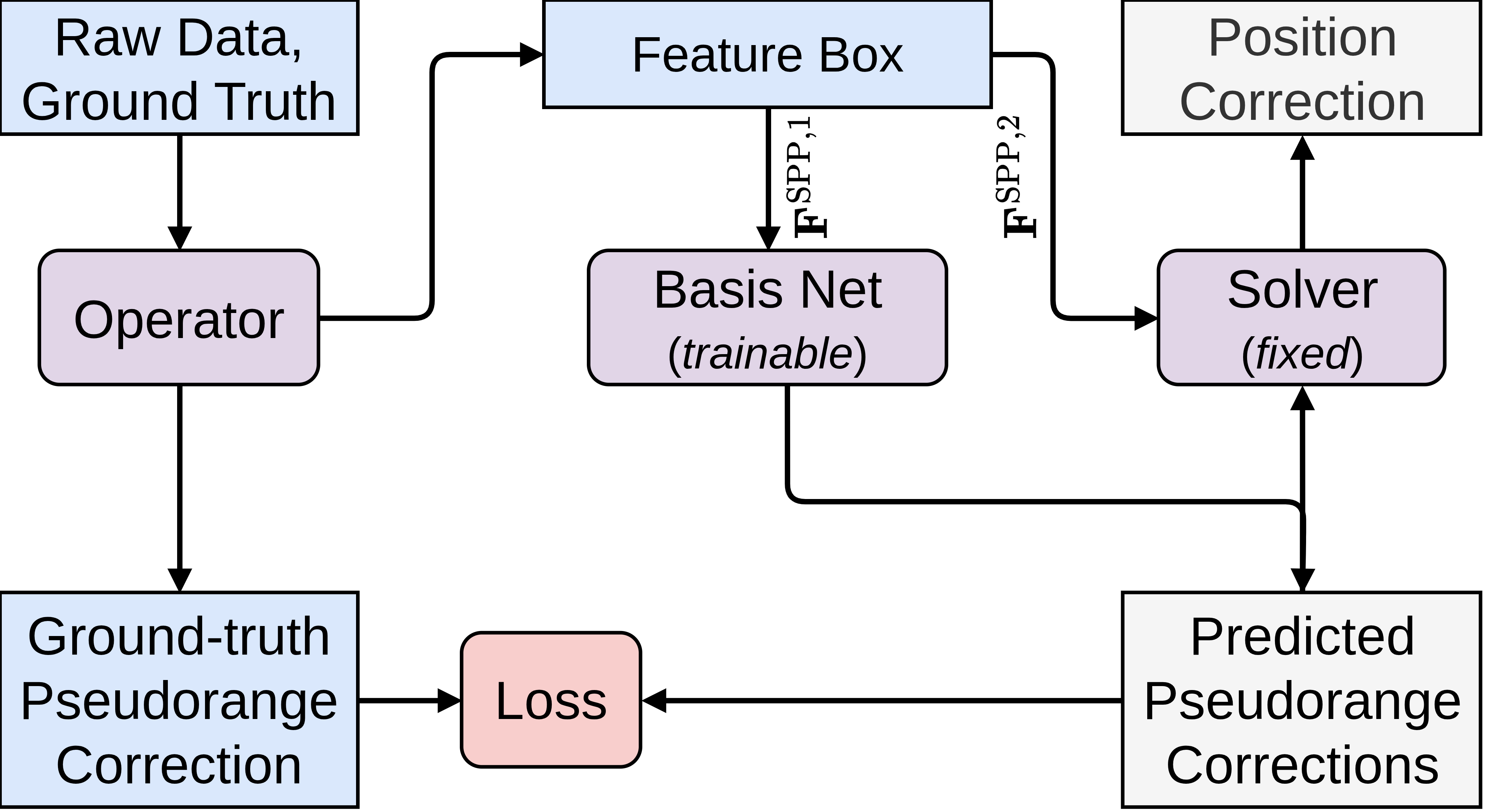}
    \caption{Training pipeline via supervised learning (SL).}
    \label{subfig:training_pipeline_SL}
\end{subfigure}

\caption{Overview of the proposed~\gnetss{} framework under the self-supervised learning (a) and supervised learning (b) schemes. For simplicity, the diagrams omit the feature-set subscripts defined in Section~\ref{sec:method}.
} 
\label{fig:training_pipeline}
\vspace{-2mm}
\end{figure*}

Other learning-based approaches directly estimate position corrections rather than correcting individual satellite measurements. \citet{Kanhere2021ImprovingCorrections} proposed a set-transformer-based architecture that takes pseudorange residuals and line-of-sight (LOS) unit vectors as input, and outputs a position correction. Their method captures the unordered set structure of satellite measurements and directly improves the GNSS position estimate. However, the training process relies on simulated multipath effects generated using random noise, which may not fully capture the structured and environment-dependent characteristics of real GNSS errors.


Overall, deep-learning-based GNSS methods have shown strong potential to model multipath, NLOS, and pseudorange errors that are difficult to describe analytically. Nevertheless, existing approaches require supervised labels from auxiliary sensors, 3D maps, or high-accuracy reference systems. In contrast, we investigate a self-supervised formulation, described in Section~\ref{sec:method}, that learns GNSS corrections without requiring ground-truth position labels.

\section{Proposed Method}
\label{sec:method}
The overall architecture of~\gnetss{} is illustrated in Figure~\ref{fig:training_pipeline}. \gnetss{} comprises three main components: an \emph{Operator}, a trainable \emph{Basis Net}, and a deterministic differentiable \emph{Solver}. 
The Operator uses raw GNSS observations and navigation messages to perform SPP, generating the features stored in the \emph{Feature Box}. 
The \emph{Basis Net} predicts satellite-wise pseudorange corrections from these features, which the \emph{Solver} converts into receiver position corrections by enforcing the GNSS measurement geometry and incorporating a dynamic model. The loss is computed from \gnetss{}'s predictions, and gradients are back-propagated through the \emph{Solver} to update the \emph{Basis Net}.

\subsection{Problem Formulation}
\label{subsec:SPP}
For satellite $s$ tracked at epoch $t$, the pseudorange (code) measurement $\rho_t^s$ is modelled as
\begin{equation}
    \rho_{t}^s = \| \mathbf{p}_{t}^s - \mathbf{p}_{r,t} \| + \eta_{t}^s, \label{eq:pseudo}
\end{equation}
where vector $\mathbf{p}_t^s$ is the position of satellite $s$, $\mathbf{p}_{r,t}$ is the true receiver position, $\|.\|$ denotes the Euclidean distance (geometric range) between them, and $\eta_t^s$ is the pseudorange measurement error which, next to random noise, also consists of modelled and unmodelled biases. In SPP, the dominant components of $\eta_t^s$ are given by
\begin{equation}
\label{eq:err_spp}
\eta_t^{s,\text{SPP}} = c \, (\Delta t_{r,t}- \Delta t_{t}^s) 
           + I_{t}^s + T_{t}^s 
           + d_t^{s,\text{Sag}},
\end{equation}
where $c$ is the speed of light, $\Delta t_{r,t}$ and $\Delta t_t^s$ are the receiver and satellite clock biases, $I_t^s$ and $T_t^s$ are the ionospheric and tropospheric delays, 
and $d_t^{s,\text{Sag}}$ denotes the Sagnac correction~\citep{Ashby2003}. These terms compensate for major systematic effects, but residual errors remain due to multipath interference, NLOS reception, receiver-dependent biases, imperfect modelling, and other environment-dependent effects. We define the remaining correction as $\epsilon_{t}^s = \eta_{t}^s - \eta_t^{s,\text{SPP}}$.

Now, let $\mathbf{p}_{r,t}^{\text{SPP}}$ be the {\em estimated version} of the receiver position by SPP. Its link to the receiver position $\mathbf{p}_{r,t}$ can be expressed as
\begin{equation}
    \mathbf{p}_{r,t} = \mathbf{p}_{r,t}^{\text{SPP}} + \Delta \mathbf{p}_{r,t},
\end{equation}
with $\Delta \mathbf{p}_{r,t}$ being the unknown receiver position correction. Linearising the geometric range around $\mathbf{p}_{r,t}^{\text{SPP}}$ gives
\begin{equation}
  \left\|
    \mathbf{p}_t^s - \mathbf{p}_{r,t}
  \right\|
  \approx 
  \left\|
    \mathbf{p}_t^s - \mathbf{p}_{r,t}^{\text{SPP}}
  \right\| - {\mathbf{u}_t^s}^\top \Delta \mathbf{p}_{r,t} ,
  \label{eq:geometric_range}
\end{equation}
where $\mathbf{u}_t^s\in\mathbb{R}^{3}$ is the LOS unit vector from the receiver to satellite $s$. The SPP pseudorange residual is computed as
\begin{equation}
  r_t^{s,\text{SPP}} =
  \rho_t^s - \left(\|
      \mathbf{p}_t^s -
      \mathbf{p}_{r,t}^{\text{SPP}}
    \| + \eta_t^{s,\text{SPP}}
  \right).
  \label{eq:res_spp}
\end{equation}

Substituting Eq.~\eqref{eq:pseudo} and Eq.~\eqref{eq:geometric_range} into Eq.~\eqref{eq:res_spp} and using the definition of $\epsilon^s_t$ yields:
\begin{equation}
    \mathbf{r}_t^{\text{SPP}} =
    -\mathbf{U}_t^{} \, \Delta \mathbf{p}_{r,t}
    + \boldsymbol{\epsilon}_t ,
    \label{eq:stacked_residual_model}
\end{equation}
where $\mathbf{r}_t^{\text{SPP}}$ stacks the SPP residuals, matrix $\mathbf{U}_t$ stacks the transposed LOS vectors $\mathbf{u}_t^s{}^\top$, and $\boldsymbol{\epsilon}_t$ contains the remaining unmodelled pseudorange errors,~\ie~$\epsilon_{t}^s$.
 Eq.~\eqref{eq:stacked_residual_model} shows that the unknown receiver correction $\Delta \mathbf{p}_{r,t}$ and unmodelled pseudorange error $\boldsymbol{\epsilon}_t$ are coupled through the known LOS geometry. Thus, learning measurement-level corrections provides information for estimating the corresponding position correction.
 Our objective is to learn these pseudorange corrections through self-supervision and use them within a differentiable solver to obtain receiver position corrections.


\subsection{Operator and Feature Box}
\label{sec:operator_feature_box}

The \emph{Operator} converts raw GNSS observations and satellite navigation messages into SPP-derived features used by both the Basis Net and the differentiable solver. 
Specifically, it computes the initial SPP receiver position $\mathbf{p}_{r,t}^{\text{SPP}}$, pseudorange residuals $r_t^{s,\text{SPP}}$, and LOS unit vectors $\mathbf{u}_t^s$, which implicitly encode satellite azimuth and elevation angles. These features are organised in the \emph{Feature Box} as satellite-wise and receiver-context feature matrices, which are used by Basis Net to predict pseudorange corrections.
At epoch $t$, for satellite $s$, the Basis-Net input feature is
\begin{equation}
    \mathbf{f}_t^s
    =
    \left[
        \mathbf{z}_t^s,
        \mathbf{c}_t
    \right],
    \label{eq:basis_feature}
\end{equation}
where $\mathbf{z}_t^s\in\mathbb{R}^{f}$ is an $f$-dimensional satellite-dependent feature vector, and $\mathbf{c}_t\in\mathbb{R}^{4}$ is a receiver-context feature vector shared by all satellites at epoch $t$. The satellite-dependent feature vector is defined as
\begin{equation}
    \mathbf{z}_t^s =
    \left[
        r_t^{s,\text{SPP}},
        \alpha_t^s,
        e_t^s,
        \text{SNR}_t^s
    \right],
    \label{eq:satellite_feature}
\end{equation}
where $\alpha_t^s$ and $e_t^s$ are the satellite azimuth and elevation angles, obtained from LOS, and $\text{SNR}_t^s$ is the ratio of the received satellite signal power to the ambient noise power, commonly expressed in dB-Hz. Note that we use $r_t^{s,\text{SPP}}$, $\alpha_t^s$ and $e_t^s$ instead of $\rho^s_t$ and $\mathbf{p}^s_t$, as they retain the essential geometric information while being more numerically stable.

The receiver-context feature $\mathbf{c}_t$ encodes the relative SPP displacement and temporal offset with respect to the current epoch, which we refer to as the \emph{target} epoch hereafter. Since Basis Net learns satellite-wise geometric relationships and temporal consistency within a sliding-window, we maintain a temporal window containing the \emph{target} epoch $t$ and $n$ \emph{historical} epochs. Historical epochs are indexed by $t-d_i,~i\in\{1,\dots, n\}$, where $d_i$ denotes the temporal offset relative to the target epoch.
For the target epoch $t$, the receiver-context feature is set to $\mathbf{c}_t=\mathbf{0}$. For a historical epoch $t-d_i$, it is defined as
\begin{equation}
    \mathbf{c}_{t-d_i}
    =
    \left[
        \mathbf{p}_{r,t-d_i}^{\text{SPP}}
        -
        \mathbf{p}_{r,t}^{\text{SPP}},
        -d_i
    \right].
    \label{eq:context_feature}
\end{equation}

We store these features in the Feature Box as the first feature block, denoted by~$\mathbf{F}_{t-d_n:t}^{\text{SPP},1}$.

The elements required by the solver are SPP pseudorange residuals and their associated weights, LOS vectors, and SPP receiver positions. Therefore, the corresponding features for satellite $s$ at epoch $t$, are $[r_t^{s,\text{SPP}}, \mathbf{u}_t^s, w_t^{s,\text{SPP}},\mathbf{p}_{r,t}^{\text{SPP}}]$, where $w_t^{s,\text{SPP}}$ denotes the pseudorange measurement weight. To incorporate dynamic models in the solver, we define a second sliding window consisting of the target epoch and $n'$ previous epochs, with the information stored in the Feature Box as the second feature tensor denoted by~$\mathbf{F}^{\text{SPP},2}_{t-d_{n'}:t}$. We make the solver window size $n'$ independent of the Basis Net window $n$. This decoupling allows the temporal context used for learning and optimisation to be configured separately, improving flexibility and computational efficiency.

When a new GNSS epoch arrives, the windows of Basis Net (described in Section~\ref{sec:basis_net}) and the Solver (described in Section~\ref{subsec:solver}) move forward with a fixed stride,~\eg~stride $1$.


\subsection{Basis Net}\label{sec:basis_net}
Inspired by prior work modelling multi-path propagation errors using basis functions~\citep{Li2019,GUO20231153}, we design \emph{Basis Net} to predict satellite-wise pseudorange corrections from SPP-derived features while maintaining real-time inference. The network estimates the pseudorange corrections $\hat{\boldsymbol{\epsilon}}_t$, which are subsequently passed to the solver to estimate the corresponding receiver position correction. This formulation keeps learning at the measurement level while preserving the geometric structure of GNSS positioning.

As described earlier, for a target epoch $t$, Basis Net takes as input a temporal window of features from the current and $n$ previous epochs encapsulated in~$\mathbf{F}^{\text{SPP,1}}_{t-d_n:t}$. Since the number of visible satellites can vary across epochs, we define $M$ as the maximum number of visible satellites and pad epochs with fewer measurements to obtain a fixed-size representation. 
During training, full trajectories are converted into overlapping windows of length $n+1$, yielding tensors of shape ($N_w, n+1, M, f+4$), where $N_w$ is the number of Basis Net windows. This formulation allows temporal windows to be processed in parallel during training.

\begin{figure}[t]
    \centering
    \includegraphics[width=0.5\linewidth]{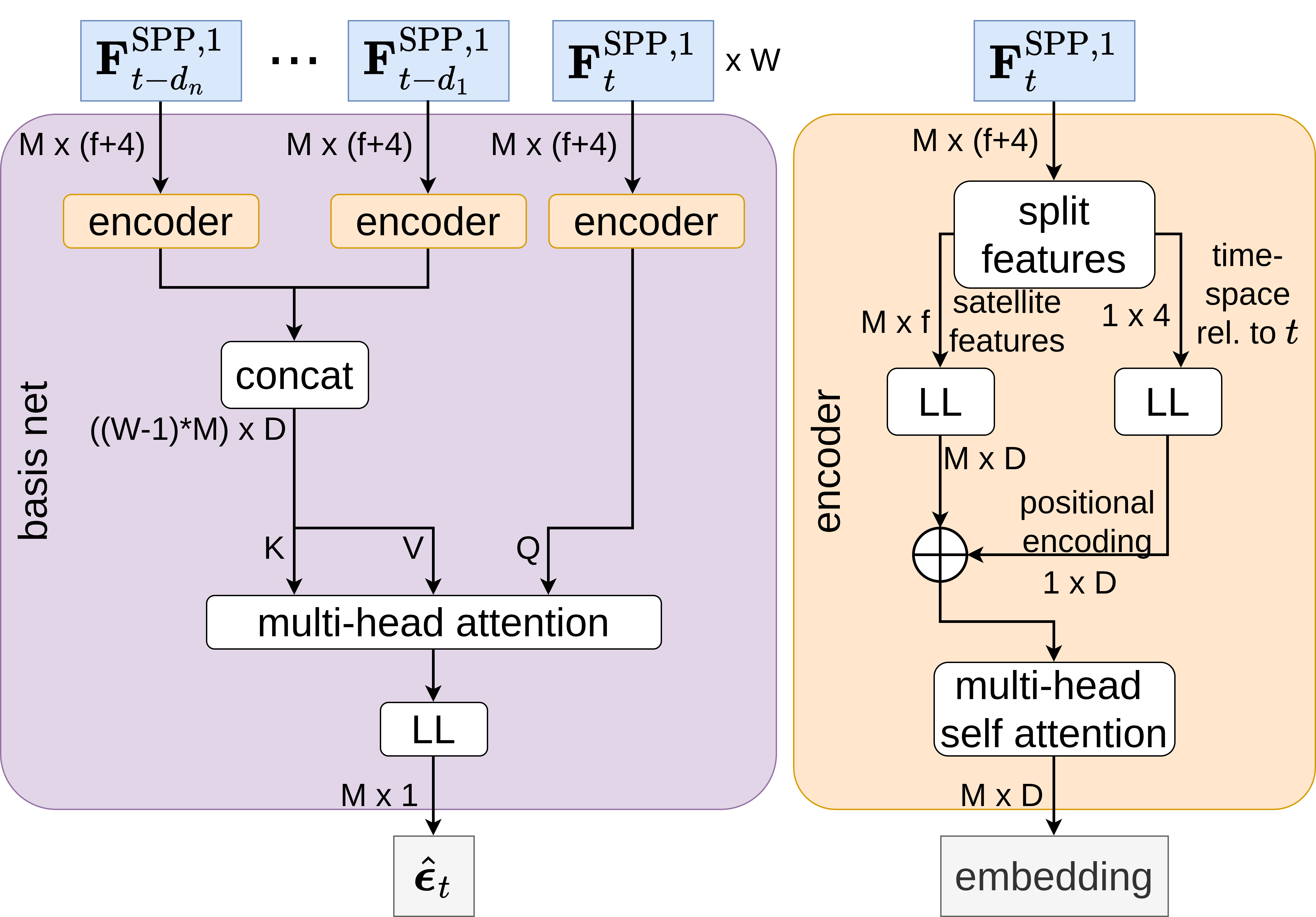}
    \caption{Basis Net architecture. \textbf{Left}: High-level temporal attention structure. 
    \textbf{Right}: Shared encoder structure. 
    }
    \label{fig:basis_cross}
    \vspace{-2mm}
\end{figure}

As illustrated in Figure~\ref{fig:basis_cross}, each epoch is first processed by a shared encoder. The encoder separately projects satellite-specific features and receiver-context features into a common $D$-dimensional latent space, fuses the resulting embeddings, and applies multi-head self-attention across satellites. This allows the network to model inter-satellite relationships 
at each epoch.

Temporal context is incorporated using a cross-attention module. The embedding of the target epoch is used as the query, while embeddings from the previous epochs within the Basis Net windows are used as keys and values. The resulting attention output captures how past satellite configurations and residual patterns relate to the current epoch. A final prediction head maps this representation to satellite-wise pseudorange corrections,
\begin{equation}
  \hat{\boldsymbol{\epsilon}}_t
  = \text{BasisNet}\left(
      \mathbf{F}^{\text{SPP},1}_{t-d_n:t}
    \right),
  \label{eq:basis_net_output}
\end{equation}
where $\hat{\boldsymbol{\epsilon}}_t \in \mathbb{R}^{M}$ contains one correction for each valid satellite observation at epoch $t$. Invalid or padded satellites are excluded by masking. 
\subsection{Differentiable Solver}
\label{subsec:solver}

Given the solver feature window $\mathbf{F}_{t-d_{N'}:t}^{\text{SPP},2}$ defined in Section~\ref{sec:operator_feature_box}, the differentiable solver estimates receiver position corrections from the pseudorange corrections predicted by the Basis Net, while enforcing temporal consistency with receiver dynamics. This is achieved through a weighted least-squares formulation that combines pseudorange residual consistency with constant-velocity regularisation. Consequently, the network learns measurement-level corrections while the final position estimate remains constrained by both LOS geometry and the assumed motion model.

For a window of $B=n'+1$ epochs, let
\begin{equation}
    \Delta {\mathbf{P}}_r
    = \left[\Delta {\mathbf{p}}_{r,1}^\top,\ldots,\Delta {\mathbf{p}}_{B}^\top\right]^\top
\label{eq:delta_p_stack}
\end{equation}
denote the stacked position corrections to be estimated by the solver. The corresponding stacked corrected receiver position is then denoted as ${\mathbf{P}}_r=[{\mathbf{p}}_{r,1}^{\top},\ldots,{\mathbf{p}}_{r,B}^{\top}]^\top$.

The solver minimises the following weighted least-squares objective
\begin{equation}
    \displaystyle
    \min_{\Delta \mathbf{P}_r}
    \sum_{t=1}^{B}
    \left\|
        \mathbf{W}_t^{1/2}
        \left(
            - \mathbf{r}_t^{\text{SPP}}
            + \hat{\boldsymbol{\epsilon}}_t
            -\mathbf{U}_t \Delta \mathbf{p}_{r,t}
        \right)
    \right\|^2
    +
    \lambda
    \left\|
        (\mathbf{D}\otimes\mathbf{I}_3)\mathbf{P}_r
    \right\|^2,  
    \label{eq:solver_objective}
\end{equation}
where~$\mathbf{W}_t = diag(w^1_t,\cdots,w^M_t)$ is the pseudorange measurement weight matrix, and hyperparameter~$\lambda$ adjusts the contribution of the regularisation. 
Symbol $\otimes$ denotes the Kronecker product. The matrix $\mathbf{D}$ encodes the second-order finite difference over valid temporal \emph{triplets} of positions, where a valid triplet consists of three consecutive epochs separated exactly by the GNSS sampling interval. Let the $k$-th valid triplet be indexed by $(i_k,i_k+1,i_k+2)$. Then,
\begin{equation}
  \mathbf{D}_{k,j}
  =
  \mathbb{I}[j=i_k]
  - 2\mathbb{I}[j=i_k+1]
  + \mathbb{I}[j=i_k+2],
  \label{eq}
\end{equation}
where $\mathbb{I}[\cdot]$ is the indicator function.

The first term in Eq.~\eqref{eq:solver_objective} enforces consistency between the pseudorange residuals and the estimated position, and the pseudorange correction. The second term imposes a constant-velocity prior over valid temporal triplets of estimated positions. When $\lambda=0$, the formulation reduces to independent epoch-wise WLS. When $\lambda>0$, it becomes a temporally regularised least-squares problem similar to MHE~\cite{4792993}. Both cases are implemented using differentiable linear-algebra operations, allowing gradients from the loss to be back-propagated through the solver to the Basis Net. During training, temporal regularisation is applied over full batches for efficient optimisation. During inference, the same formulation is applied over a fixed window containing the current and past states.


\subsection{Loss Formulation}
\label{subsec:loss}
The proposed \gnetss{} framework is trained using a self-supervised objective. The key idea is to predict pseudorange corrections such that the corrected measurements become consistent with the underlying GNSS geometry and receiver motion. To this end, we leverage two complementary spatio-temporal constraints: trajectory-level velocity consistency and measurement-level residual reconstruction. 

At the trajectory level, Doppler observations are used to estimate the receiver velocity at each epoch, denoted by $\mathbf{v}_{r,t}^{\text{SPP}}$. Since Doppler measurements correspond to pseudorange rates rather than absolute pseudoranges, they are generally less affected by slowly varying measurement biases and thus provide a better estimate of instantaneous velocity ~\cite{enge1994global}. The predicted velocity is computed from consecutive position estimates as $\hat{\mathbf{v}}_{r,t} = \big(\hat{\mathbf{p}}_{r,t}-\hat{\mathbf{p}}_{r,t-\Delta t}\big)/ \Delta t$, where $\Delta t$ is the GNSS sampling interval. We therefore define a velocity consistency loss $\mathcal{L}_v$ as
\begin{equation}
    \mathcal{L}_v
    =
    \frac{1}{N_v}
    \sum_{t=1}^{N_v}
    \ell_{\text{Huber}}
    \left(
    \mathbf{v}_{r,t}^{\text{SPP}},
    \hat{\mathbf{v}}_{r,t}
    \right),
    \label{eq:velocity_loss}
\end{equation}
where $N_v$ is the number of velocity measurements, corresponding to position estimate pairs separated exactly by $\Delta t$, and $\ell_{\text{Huber}}$ denotes the Huber loss. 

At the measurement level, we encourage the predicted corrections to explain the residual structure observed in the SPP solution. Using Eq.~\eqref{eq:stacked_residual_model}, a reconstructed residual $\hat{r}_t^s$ can be computed from $\hat{\epsilon}_t^s$ and $\Delta\hat{\mathbf{p}}_{r,t}$, before comparing against the corresponding SPP residual $r_t^{s,\text{SPP}}$. While the discrepancy between $\hat{r}_t^s$ and $r_t^{s,\text{SPP}}$ already appears in the optimisation objective of the solver (Eq.~\eqref{eq:solver_objective}), introducing an explicit reconstruction loss directly encourages the learned corrections to remain compatible with the underlying GNSS measurement geometry. Using the $L_1$ norm $|\cdot|$, we define a weighted mean absolute error (WMAE) residual reconstruction loss $\mathcal{L}_r$ as
\begin{equation}
    \mathcal{L}_r
    = \frac{1}{W}
        \sum_{t=1}^{N}
        \sum_{s=1}^{M_t}
        w_t^{s,\text{SPP}}\,
        \left|\hat{r}^s_t - r^{s,\text{SPP}}_t\right|,
    W = \sum_{t=1}^{N} \sum_{s=1}^{M_t} w_t^{s,\text{SPP}},
\label{eq:residual_consistency_loss}
\end{equation}
where $M_t$ is the number of valid satellites at epoch $t$. The final self-supervised learning (SSL) loss is:
\begin{equation}
    \mathcal{L}_{\text{SSL}}
    = \lambda_r \mathcal{L}_r + \lambda_v \mathcal{L}_v ,
\label{eq:self_supervised_loss}
\end{equation}
where $\lambda_r$ and $\lambda_v$ control the contribution of the residual and velocity consistency, respectively.

To assess the effectiveness of the proposed self-supervised training strategy, we additionally train \gnetss{} using full supervision. Specifically, the network is trained with supervision on $\epsilon_t^s$. Given ground-truth (GT) receiver positions, the true position correction relative to the SPP solution is obtained as $\Delta\mathbf{p}_{r,t}^{\text{GT}}=\mathbf{p}_{r,t}^{\text{GT}}-\mathbf{p}_{r,t}^{\text{SPP}}$. Substituting $\Delta\mathbf{p}_{r,t}^{\text{GT}}$ into Eq.~\eqref{eq:stacked_residual_model} yields the corresponding target correction $\epsilon_t^{s,\text{SPP}}$. Basis Net is then trained using a weighted mean squared error (WMSE) objective:
\begin{equation}
    \mathcal{L}_{\text{SL}}
    =
    \frac{1}{W}
    \sum_{t=1}^{N}
    \sum_{s=1}^{M_t}
    w_t^{s,\text{SPP}}
    \left\|
        \hat{\epsilon}_t^s
        -
        \epsilon_t^{s,\text{SPP}}
    \right\|^2.
\label{eq:supervised_loss}
\end{equation}

\section{Experiments}\label{sec:experiments}
This section presents the experimental design and evaluation of GNetSS. We begin by describing the public and in-house datasets used to train and evaluate GNetSS, followed by the implementation details, including the training configuration and evaluation metrics. Next, we introduce the baseline methods used for comparison and present the comparatative results. Finally, an ablation study is conducted to assess the contribution of temporal cross-attention and dynamic model, as well as the effect of their corresponding window size on positioning accuracy.

\subsection{Datasets}
We train and evaluate~\gnetss{} on three datasets: Google Smartphone Decimeter Challenge (GSDC)~\citep{fu2021google_smartphone_decimeter}, UrbanNav~\citep{Hsunavi} and an \datasetname{} dataset. All datasets have a sampling rate of $\SI{1}{\Hz}$, and are summarised in Table~\ref{tab:datasets}. We restrict the experiments to GPS L1 observations, providing a controlled single-constellation, single-frequency setting for evaluating the proposed self-supervised framework, leaving multi-constellation and multi-frequency extensions for future work. Epochs with fewer than four valid satellite measurements are discarded.
For GSDC, we observed that the derived ground-truth files contain an error in the conversion from orthometric to ellipsoidal height, with geoid separation subtracted rather than added. Thus, we recomputed ellipsoidal heights from the raw ground-truth files. We also excluded Xiaomi8 data due to detecting unreliable Doppler measurements.

Following the standard train-validation-test paradigm described by~\citep{Hastie2009}, we split the filtered data from each dataset into training, validation, and evaluation sets in an approximate \(7{:}1{:}2\) ratio.
While most DL methods (\eg~\citep{Kanhere2021ImprovingCorrections}) perform a random timestamp-device level split, the temporal dependencies in \gnetss{} require all samples from a trajectory to remain in the same partition. Thus, we group trajectories by collection sessions, and sessions are manually assigned to the validation and evaluation partitions such that they sum to approximately $10\%$ and $20\%$ of the data, respectively, with no session appearing in multiple partitions.

\begin{figure}[bt] 
    \centering
    \begin{subfigure}[b]{0.33\linewidth}
        \centering
        \includegraphics[width=0.95\linewidth]{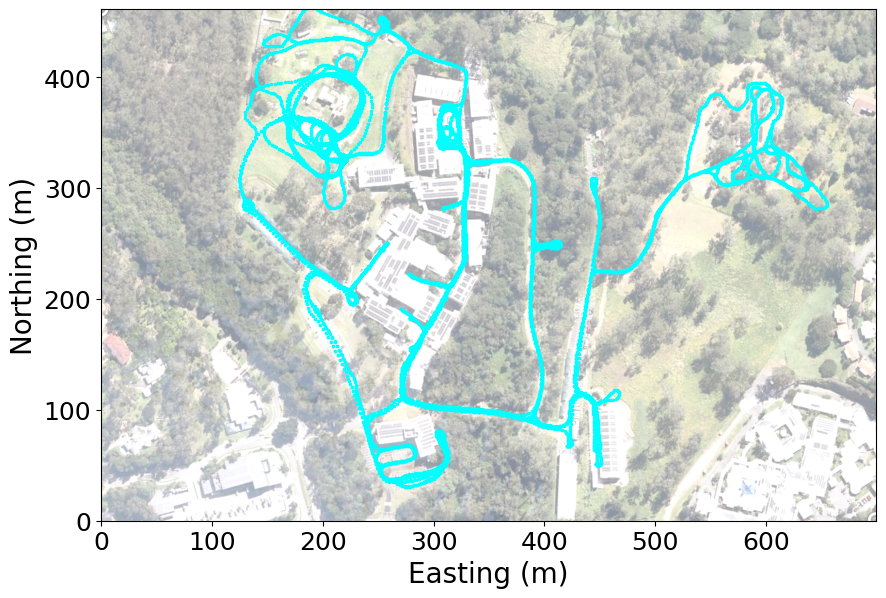}
    \end{subfigure}
    \hfill
    \begin{subfigure}[b]{0.33\linewidth}
        \centering
        \includegraphics[width=0.95\linewidth]{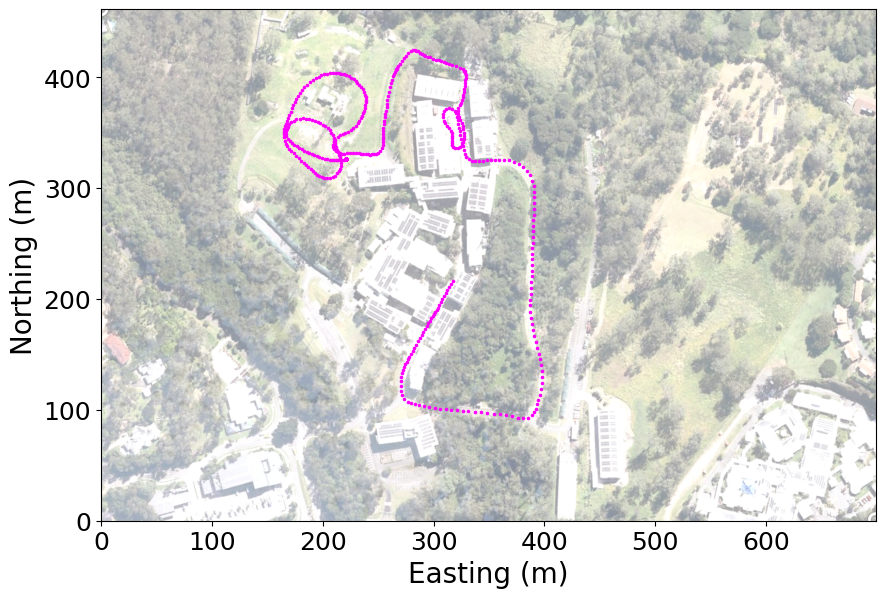}
    \end{subfigure}
    \hfill
    \begin{subfigure}[b]{0.33\linewidth}
        \centering
        \includegraphics[width=0.95\linewidth]{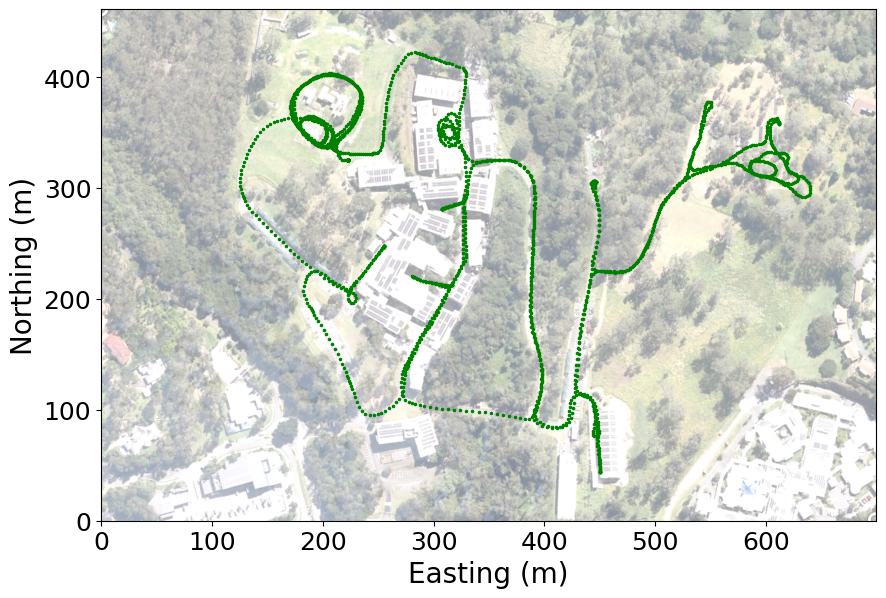}
    \end{subfigure}
    \caption{Training (left), validation (middle), and evaluation (right) set for \datasetname{} dataset.}
    \label{fig:CSIRO_split} 
    \vspace{-2mm}
\end{figure}

\begin{figure}[bt] 
    \centering
    \begin{subfigure}[b]{0.33\linewidth}
        \centering
        \includegraphics[width=0.95\linewidth]{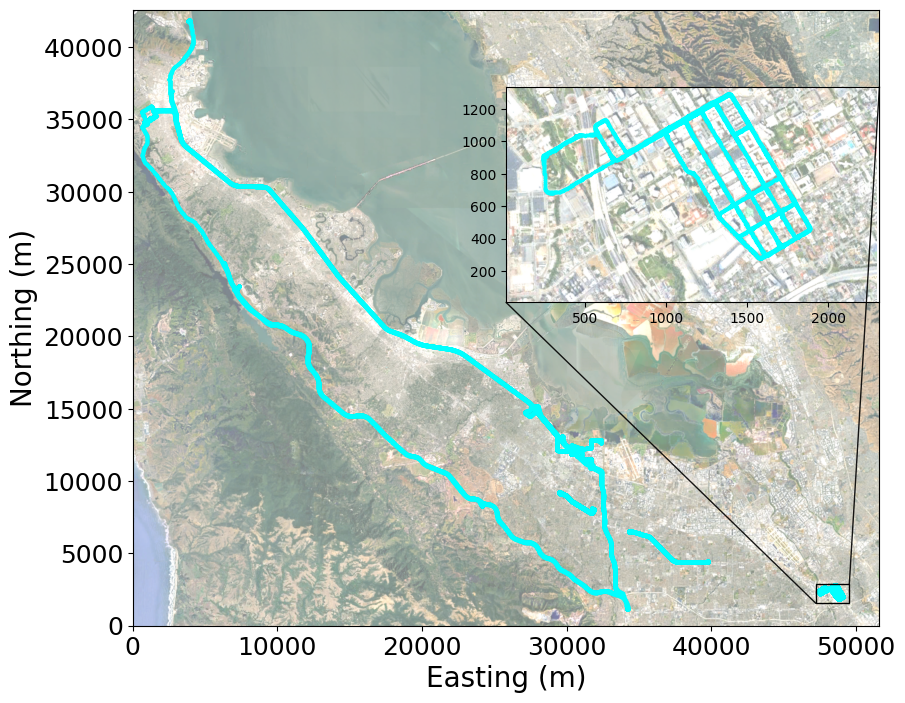}
    \end{subfigure}
    \hfill
    \begin{subfigure}[b]{0.33\linewidth}
        \centering
        \includegraphics[width=0.95\linewidth]{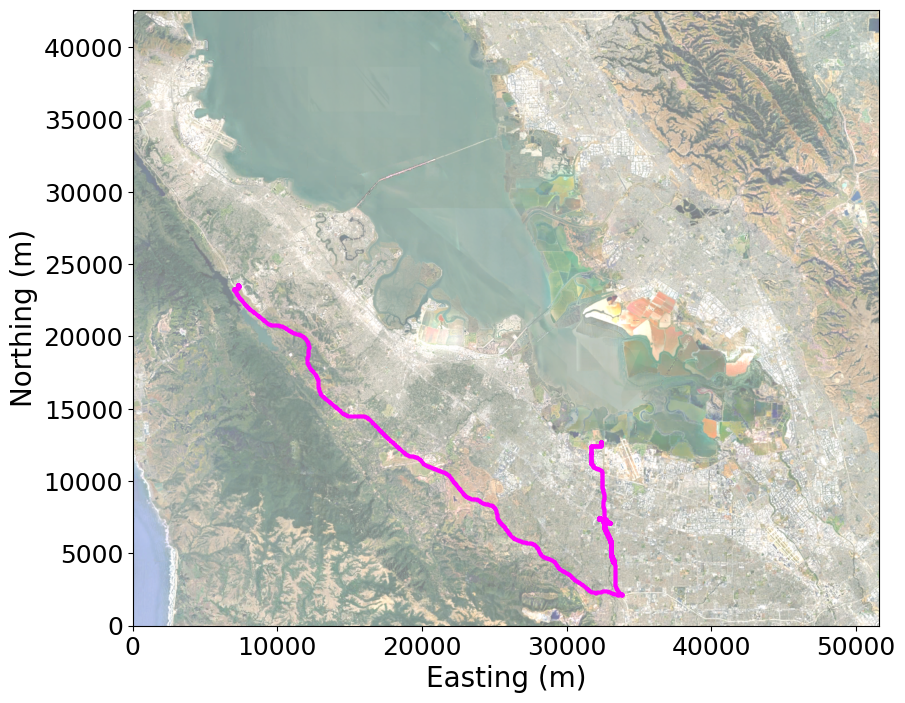}
    \end{subfigure}
    \hfill
    \begin{subfigure}[b]{0.33\linewidth}
        \centering
        \includegraphics[width=0.95\linewidth]{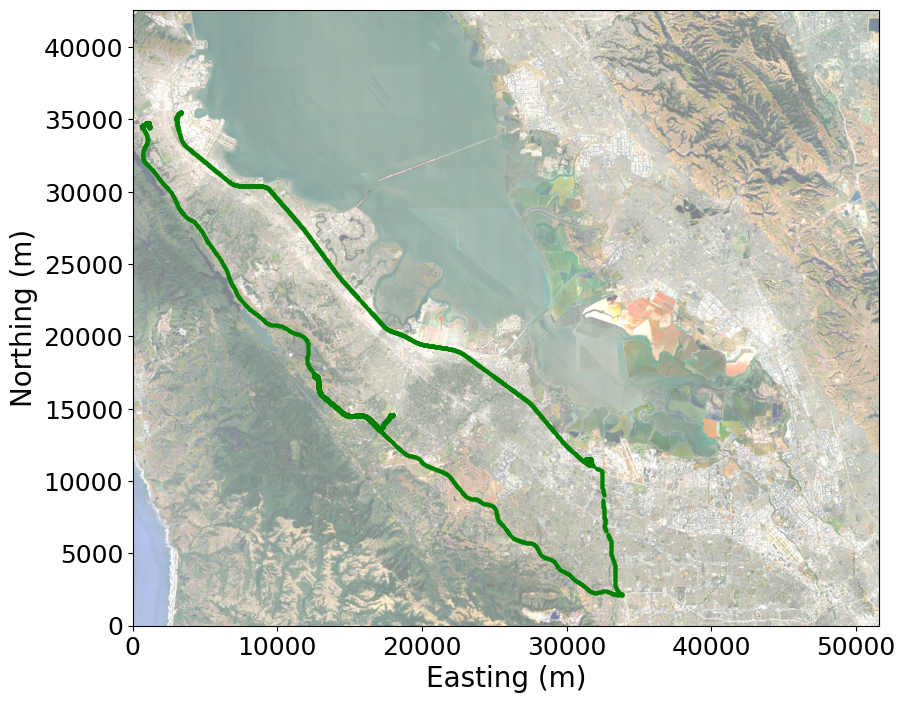}
    \end{subfigure}
    \caption{Training (left), validation (middle), and evaluation (right) set for one of the partitioning configuration of GSDC dataset.}
    \label{fig:GSDC_split} 
    \vspace{-2mm}
\end{figure}

\begin{figure}[bt] 
    \centering
    \begin{subfigure}[b]{0.33\linewidth}
        \centering
        \includegraphics[width=0.75\linewidth]{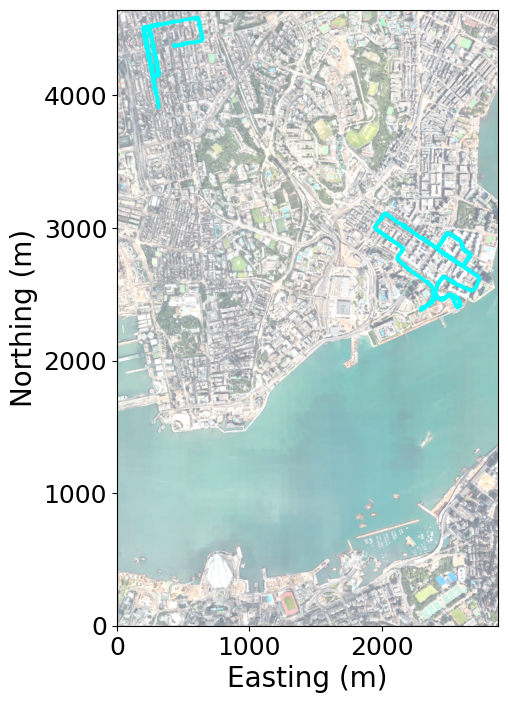}
    \end{subfigure}
    \hfill
    \begin{subfigure}[b]{0.33\linewidth}
        \centering
        \includegraphics[width=0.75\linewidth]{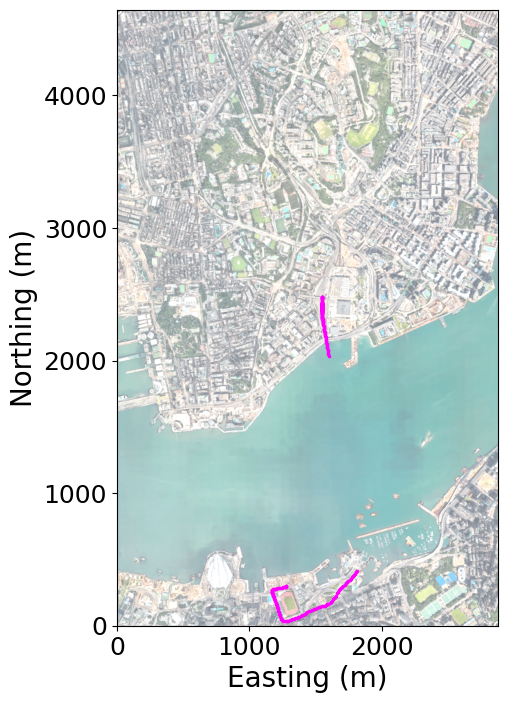}
    \end{subfigure}
    \hfill
    \begin{subfigure}[b]{0.33\linewidth}
        \centering
        \includegraphics[width=0.75\linewidth]{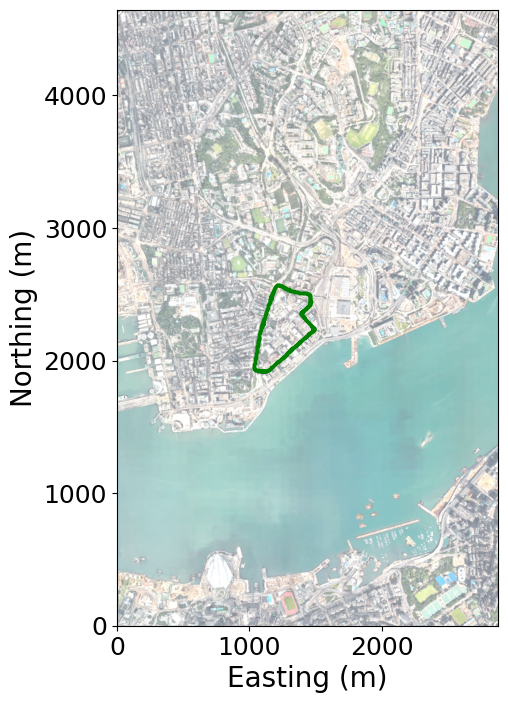}
    \end{subfigure}
    \caption{Training (left), validation (middle), and evaluation (right) set for UrbanNav dataset.}
    \label{fig:UrbanNav_split} 
    \vspace{-2mm}
\end{figure}

Due to the small number of collection sessions in UrbanNav and \datasetname{} dataset, there is only one possible partition configuration close to the target  \(7{:}1{:}2\) ratio for each dataset. However, GSDC has higher number of collection sessions, allowing for more partition configuration. In this work, 10 configurations were selected. Results on GSDC are reported as aggregates across all 10 partitions to reduce sensitivity to a particular partition configuration. Figures \ref{fig:CSIRO_split}--\ref{fig:UrbanNav_split} illustrate the partition configurations for each dataset, overlaid on the corresponding satellite imagery.


\begin{table*}[t] \caption{Summary of datasets used in this work.} \label{tab:datasets} \begin{tblr}{ width=\textwidth, colspec={ X[1.2,l] X[1.4,l] X[2.8,l] X[1.5,l] X[1,c] X[1,c] }, row{odd}={bg=black!10,font=\footnotesize}, row{even}={white,font=\footnotesize}, row{1}={bg=black!20,font=\bfseries\footnotesize}, hline{Z}={1pt,solid,black!60}, rowsep=3pt } Dataset & Ground Truth & GNSS Devices & Environment & Distance traversed & Samples \\ GSDC~\citep{fu2021google_smartphone_decimeter,fu2020android_raw_gnss} & NovAtel SPAN & Pixel4, Pixel4 XL, Pixel5, Samsung S20 Ultra & Highway, urban & \SI{928.8}{\kilo\metre} & 112,125 \\ UrbanNav~\citep{Hsunavi} & NovAtel SPAN-CPT+IE & Pixel4, Huawei P40 Pro, Samsung Note8, Xiaomi8, u-blox EVK-M8T, u-blox ZED-F9P & Dense and high-rise urban & \SI{26.9}{\kilo\metre} & 18,405 \\ \datasetname{} & LiDAR-IMU SLAM + RTK & u-blox ZED-F9P & Open-sky, forested, urban & \SI{14.4}{\kilo\metre} & 11,262 \\ \end{tblr} \end{table*}

\subsection{Implementation}
\textbf{Experimental Setup}:
\gnetss{} is implemented in \texttt{PyTorch}, and trained on a single NVIDIA H100 GPU. We used the \texttt{AdamW} optimiser with weight decay of $10^{-4}$. A validation-driven learning rate scheduler~\citep{darken1990note} is employed, initialised at $10^{-3}$ with factor $0.7$ and patience $10$. Early stopping is triggered when the validation loss does not improve for $20$ consecutive training iterations~\footnote{A single iteration is what would traditionally be referred to as a single training epoch, here we avoid the term epoch to avoid confusion with GNSS terminology}. 
For the SSL loss, we set~$\lambda_v = 5$ and $\lambda_r =1$. The Basis Net window size is set to $10$. All multi-head attention modules use a hidden dimension of $D=64$ with $4$ attention heads. Basis Net contains one self-attention layer and one cross-attention layer. During inference, the solver window size is set to $50$.

\textbf{Evaluation Metrics}:
We evaluate positioning performance using 3D position error and its 2D horizontal component. 
Given the predicted position correction $\Delta \hat{\mathbf{p}}_{r,t}$ and the GT position correction $\Delta\mathbf{p}^{\text{GT}}_{r,t}$,
\begin{align}
    \text{3D error} &= \| \Delta \hat{\mathbf{p}}_{r,t} - \Delta {\mathbf{p}}^{\text{GT}}_{r,t} \|_2, \\
    \text{2D error} &= \| (\Delta \hat{\mathbf{p}}_{r,t} - \Delta {\mathbf{p}}^{\text{GT}}_{r,t})_{1:2} \|_2\nonumber .
\end{align}
For each metric, we plot the empirical cumulative distribution function (ECDF) to compare performance across error thresholds. We also report the mean and maximum errors to summarise average and worst-case performance, respectively.



\subsection{Baseline Methods}
We compare~\gnetss{} against four baselines: SPP, EKF, MHE, and NavLab. 
SPP is implemented using RTKLIB~\citep{takasu2009development} and provides the initial receiver positions used to derive the input features for both \gnetss{} and NavLab, making it the primary reference baseline for evaluating positioning improvements. EKF and MHE follow the implementation of~\citep{weng2023localization}; both perform dynamic-model-based state estimation and are closely related to the deterministic solver in \gnetss{}, but unlike \gnetss{}, they rely on predefined measurement, motion, and noise models rather than learning measurement-level pseudorange corrections.

NavLab~\citep{Kanhere2021ImprovingCorrections} requires an initial position estimate. In the original formulation, this estimate is generated by perturbing the ground-truth position and the same procedure is used during inference. Since ground truth is unavailable in practical deployment, we instead initialise NavLab with the SPP solution at inference time. Due to its reported sensitivity to initialisation~\citep{Kanhere2021ImprovingCorrections}, we train NavLab under two settings. The first uses the SPP solution for initialisation during training, thereby matching the inference setting. The second, denoted by~$\dagger$, follows the original protocol, with the perturbation scale $\mu$ set to the $95^{\mathrm{th}}$ percentile of the SPP position error. 

\subsection{Comparative Results}\label{subsec:results}
\begin{figure}[bt] 
    \centering
    \begin{subfigure}[b]{0.33\linewidth}
        \centering
        \includegraphics[width=0.95\linewidth]{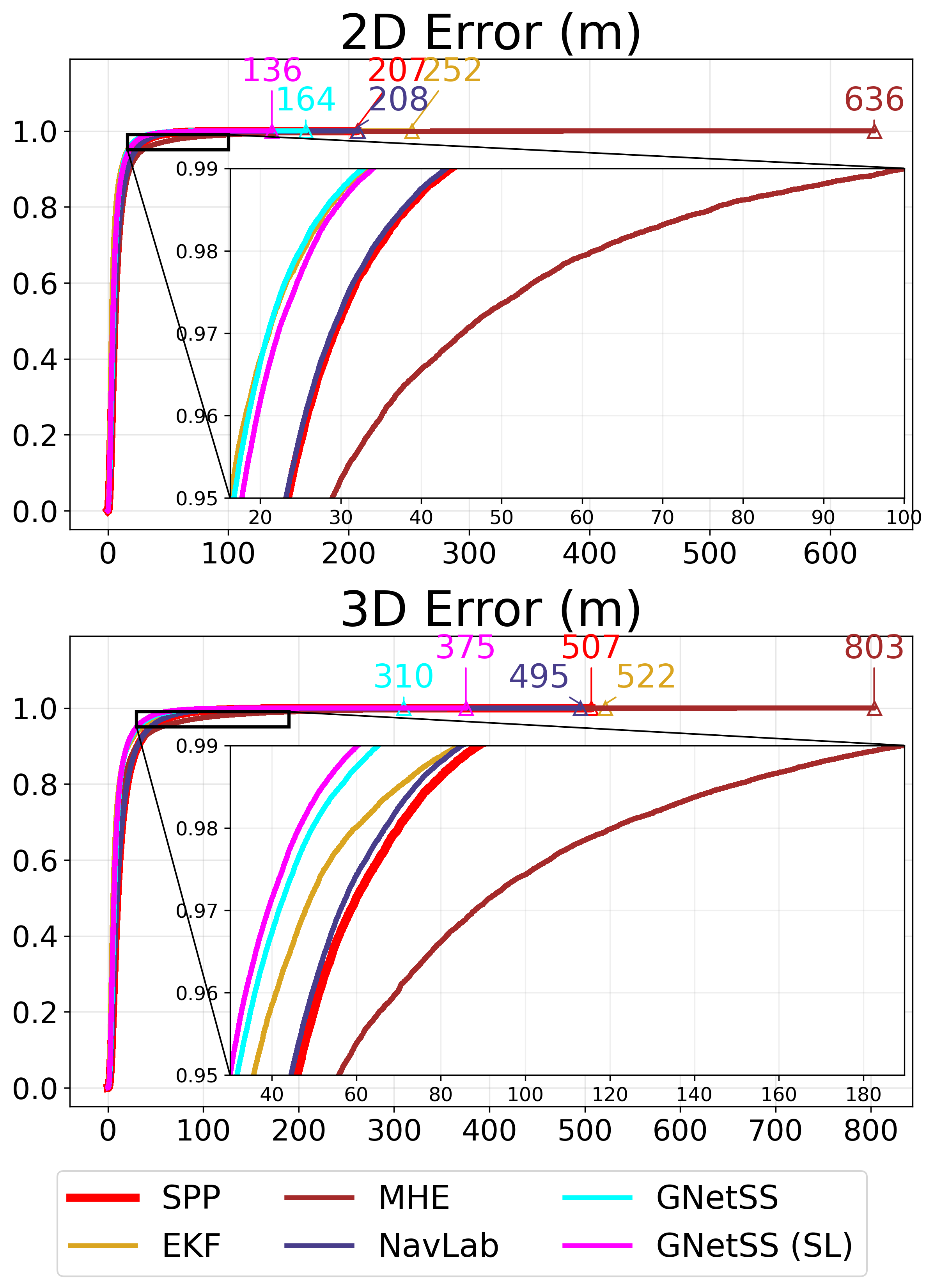}
        \caption{GSDC}
    \end{subfigure}
    \hfill
    \begin{subfigure}[b]{0.33\linewidth}
        \centering
        \includegraphics[width=0.95\linewidth]{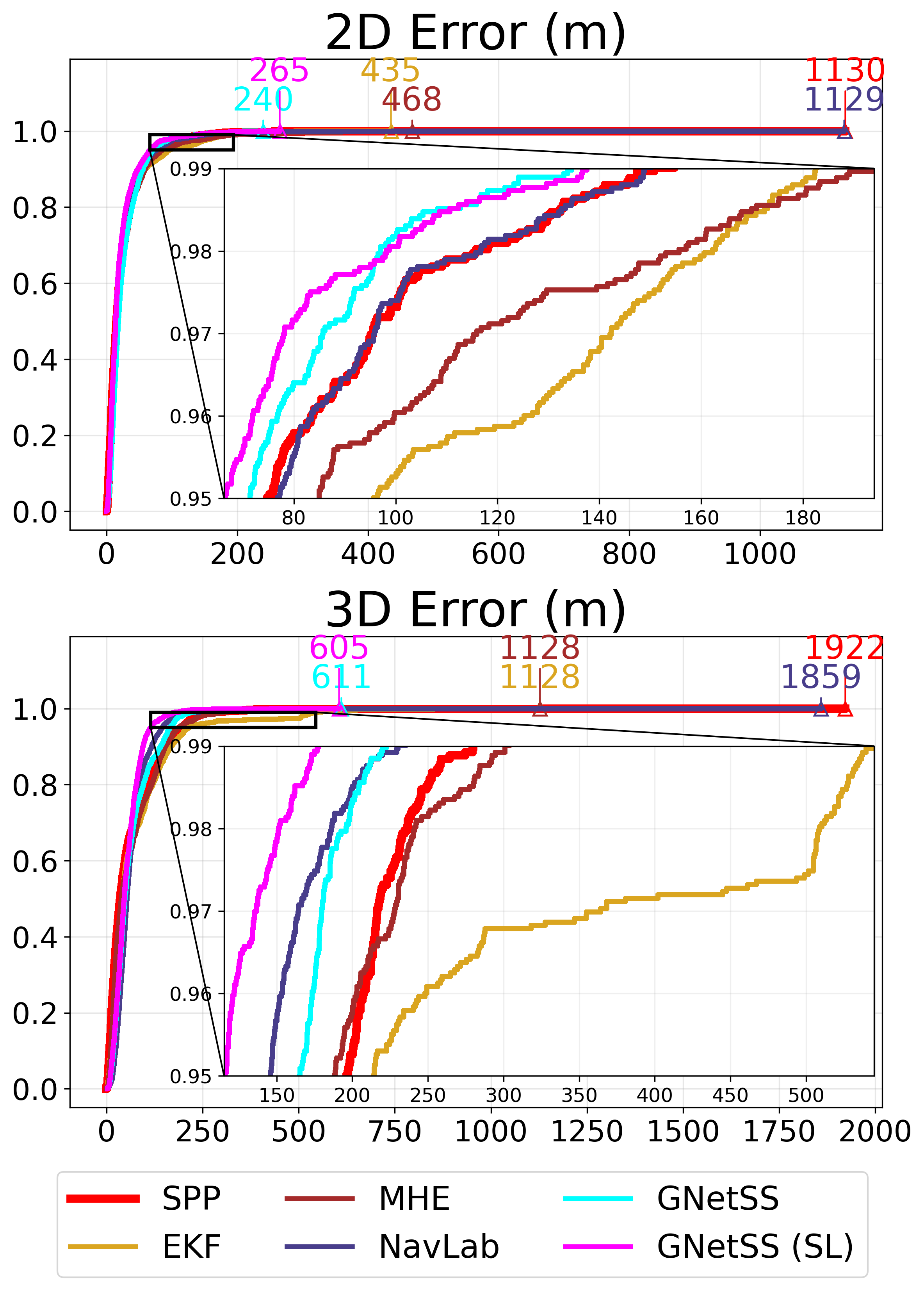}
        \caption{UrbanNav}
    \end{subfigure}
    \hfill
    \begin{subfigure}[b]{0.33\linewidth}
        \centering
        \includegraphics[width=0.95\linewidth]{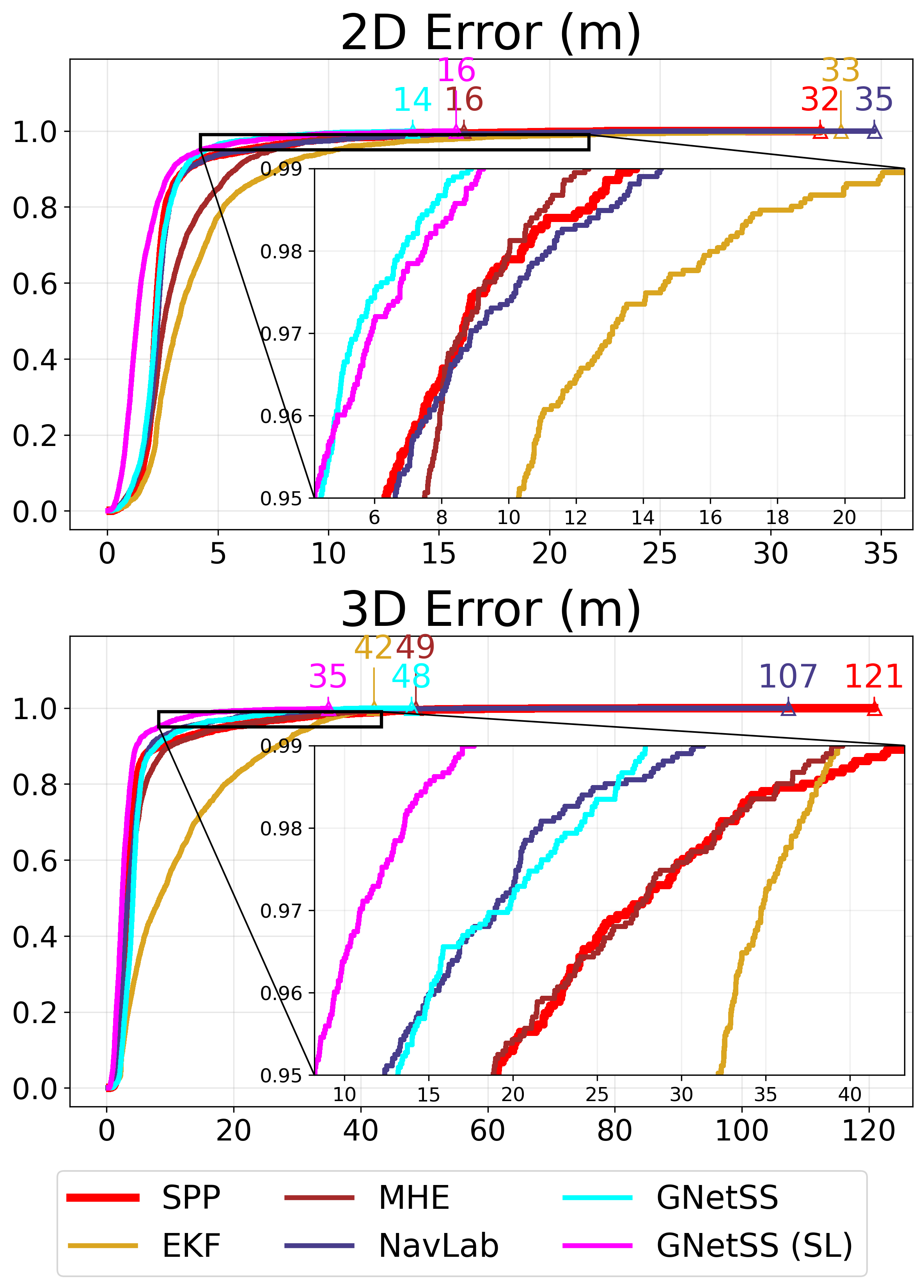}
        \caption{\datasetname{}}
    \end{subfigure}
    \caption{ECDF for GSDC (left), UrbanNav (middle) and \datasetname{} (right) datasets.}
    \label{fig:cdfs} 
    \vspace{-2mm}
\end{figure}

Figure~\ref{fig:cdfs} shows ECDFs of the 2D and 3D positioning errors on the GSDC, UrbanNav, and \datasetname{} datasets. Table~\ref{tab:performance} summarises the mean and maximum 2D and 3D positioning errors across the three datasets. 

As seen in the zoomed regions of Figure~\ref{fig:cdfs}, \gnetss{} improves the tail behaviour of the error distribution, achieving lower errors at high-percentiles than SPP, MHE, and NavLab in most cases. Compared with SPP, \gnetss{} reduces the 99th-percentile 3D error by $\sim\SI{25}{\metre}$, $\sim\SI{60}{\metre}$, and $\sim\SI{15}{\metre}$ on the GSDC, UrbanNav, and \datasetname{} datasets, respectively. This improvement is also reflected in the maximum-error statistics in Table~\ref{tab:performance}, which are consistent with the tail behaviour of Figure~\ref{fig:cdfs}, demonstrating stronger suppression of large positioning outliers. The largest gains are observed on GSDC and UrbanNav, where the maximum 3D error is reduced by $\SI{196.7}{\metre}$ and $\SI{1310.8}{\metre}$, respectively, compared to SPP. We focus on these high-percentile and maximum-error metrics because large positioning failures strongly affect system reliability.

\begin{table*}[t] \caption{Performance evaluation on individual and combined dataset} \label{tab:performance} \begin{threeparttable} \begin{tblr}{ 
    width=\textwidth, colspec={ X[3,l] *{12}{X[1,c]} *{2}{X[2,c]} }, row{odd}={bg=black!10,font=\footnotesize}, row{even}={white,font=\footnotesize}, row{1-3}={bg=black!20,font=\bfseries\footnotesize}, hline{Z}={1pt,solid,black!60}, hline{7}={0.6pt,solid,black!40}, hline{9}={0.6pt,solid,black!40}, rowsep=2pt 
    } 
    \SetCell[r=3]{l} Method & \SetCell[c=4]{c} GSDC & & & & \SetCell[c=4]{c} UrbanNav & & & & \SetCell[c=4]{c} \datasetname{} & & & & \SetCell[c=2]{c} Average & \\ 
    & \SetCell[c=2]{c} 2D error (m) && \SetCell[c=2]{c} 3D error (m) && \SetCell[c=2]{c} 2D error (m) && \SetCell[c=2]{c} 3D error (m) && \SetCell[c=2]{c} 2D error (m) && \SetCell[c=2]{c} 3D error (m) && 2D error (m) & 3D error (m) \\ 
    & Mean & Max & Mean & Max & Mean & Max & Mean & Max & Mean & Max & Mean & Max & Mean & Mean \\ SPP~\citep{takasu2009development} & 7.9 & 207.0 & 15.1 & 506.8 & \underline{24.4} & 1130.3 & \underline{62.2} & 1921.8 & 2.7 & 32.2 & 5.1 & 120.8 & 11.7 & 27.5 \\ 
    EKF~\citep{weng2023localization} & \textbf{5.1} & 252.4 & \underline{10.3} & 521.5 & 27.3 & 435.2 & 77.2 & 1127.6 & 4.3 & 33.2 & 11.5 & \underline{42.1} & 12.2 & 33.0 \\ 
    MHE~\citep{weng2023localization} & 10.4 & 636.5 & 17.4 & 803.3 & 28.9 & 467.7 & 66.7 & 1127.6 & 3.2 & 16.1 & 5.6 & 48.7 & 14.2 & 29.9 \\ 
    NavLab~\citep{Kanhere2021ImprovingCorrections} & 7.8 & 207.6 & 14.9 & 495.1 & 25.5 & 1128.6 & 64.9 & 1858.6 & 2.7 & 34.7 & \underline{4.7} & 107.3 & 12.0 & 28.2 \\ 
    NavLab\tnote{$\dagger$}~\citep{Kanhere2021ImprovingCorrections} & 9.1 & 205.3 & 17.7 & 481.4 & 24.5 & 1130.7 & 71.0 & 1871.6 & 2.5 & 32.5 & 8.8 & 114.6 & 12.0 & 32.5 \\ 
    GNetSS (SSL) & 6.1 & \underline{164.3} & 10.4 & \textbf{310.1} & 26.2 & \textbf{239.8} & 63.8 & \textbf{611.0} & \underline{2.4} & \textbf{13.8} & 5.2 & 48.0 & \underline{11.6} & \underline{26.5} \\ 
    GNetSS (SL) & \underline{6.0} & \textbf{136.2} & \textbf{9.8} & \underline{375.3} & \textbf{21.7} & \underline{265.0} & \textbf{53.8} & \underline{605.0} & \textbf{1.8} & \underline{15.8} & \textbf{3.2} & \textbf{34.9} & \textbf{9.8} & \textbf{22.3} \\ 
    \end{tblr} 
    \begin{tablenotes}[flushleft] \footnotesize \item[$\dagger$] Method is trained with random initial position guesses, while testing is initialised from SPP. \end{tablenotes} 
    \end{threeparttable} \end{table*}

When averaged across all datasets, both \gnetss{} variants outperform the baselines in terms of mean positioning accuracy, with supervised \gnetss{} achieving the strongest overall performance. The self-supervised variant remains competitive across datasets, despite not using ground-truth positions during training, indicating that the proposed strategy is effective across different GNSS conditions. In contrast, model-based baselines exhibit stronger dependencies on data characteristics, with their relative rankings varying substantially across datasets.
For example, on UrbanNav and \datasetname{}, EKF performs worse than SPP. We hypothesise that this is due to inaccurate system and noise models, particularly the assumption of zero-mean measurement noise in the presence of large outliers caused by multipath and NLOS effects. In contrast,
\gnetss{} mitigates biased pseudorange measurements before position estimation through learned pseudorange corrections, while still allowing the solver to leverage the dynamic model.
Similarly, NavLab$^{\dagger}$ performs worse than NavLab in most cases, indicating sensitivity to the initialisation strategy. This behaviour is also reflected in the ECDFs, where the NavLab curves often closely follow those of SPP, particularly in the 2D case.
These results suggest that methods relying strongly on initial position estimates can struggle when the initialisation differs between training and deployment. 

\subsection{Ablation Study}
\textbf{Cross-Attention:}
We evaluate the contribution of temporal cross-attention and the effect of the Basis Net window size on positioning accuracy on GSDC. Four configurations are considered: cross-attention disabled, corresponding to a window size of $1$, and cross-attention enabled with window sizes of $5$, $10$, and $20$. 
Table~\ref{tab:win_ablation} shows that incorporating temporal context through cross-attention improves overall performance compared with the single-epoch configuration, suggesting that temporal context is beneficial for learning pseudorange corrections in the SSL setting, as neighbouring epochs provide additional information about persistent residual patterns and receiver motion. 
However, a window size beyong $10$ provides diminishing improvement despite requiring more computing resources. A window size of $10$ therefore provides a good balance between correction accuracy and computational overhead.


\textbf{Dynamic Model:}
The second ablation study evaluates the contribution of the dynamic model and the effect of solver window size on positioning accuracy on GSDC. As shown in Table~\ref{tab:hist_ablation}, disabling the dynamic model, corresponding to a solver window size of $1$, results in higher errors. 
Enabling the dynamic model with a solver window size of $50$ reduces the 2D and 3D maximum errors to $\SI{164.3}{\metre}$ and $\SI{310.1}{\metre}$, respectively. Increasing the solver window size to $100$ provides no further improvement, indicating that a moderate temporal window is sufficient to capture the benefits of the dynamic model while avoiding unnecessary computational overhead.


\begin{table}[t] \caption{Network window size ablation study results} \label{tab:win_ablation} 
\begin{tblr}{ 
    colspec={X[l]X[c]X[c]X[c]X[c]}, width=\textwidth, row{odd}={bg=black!10,font=\footnotesize}, row{even}={white,font=\footnotesize}, row{1-2}={bg=black!20,font=\bfseries\footnotesize}, hline{Z}={1pt,solid,black!60}, rowsep=1pt } 
    \SetCell[r=2]{l} Network window size & \SetCell[c=2]{c} 2D error (m) & & \SetCell[c=2]{c} 3D error (m) & \\ 
    & Mean & Max & Mean & Max \\ 
    1 & 6.4 & 172.4 & 12.4 & 459.3 \\ 
    5 & 6.1 & 162.4 & 10.6 & 404.7 \\ 
    10 & 6.1 & 164.3 & 10.4 & 310.1 \\ 
    20 & 6.0 & 161.9 & 10.5 & 301.3 \\ 
\end{tblr} \end{table}

\begin{table}[t] \caption{Solver window size ablation study results} \label{tab:hist_ablation} 
\begin{tblr}{ 
    colspec={X[l]X[c]X[c]X[c]X[c]}, width=\textwidth, row{odd}={bg=black!10,font=\footnotesize}, row{even}={white,font=\footnotesize}, row{1-2}={bg=black!20,font=\bfseries\footnotesize}, hline{Z}={1pt,solid,black!60}, rowsep=1pt 
    } 
    \SetCell[r=2]{l} Solver window size & \SetCell[c=2]{c} 2D error (m) & & \SetCell[c=2]{c} 3D error (m) & \\ 
    & Mean & Max & Mean & Max \\ 
    1 & 6.0 & 175.8 & 10.4 & 348.8 \\ 
    50 & 6.1 & 164.3 & 10.4 & 310.1 \\ 
    100 & 6.1 & 164.3 & 10.4 & 310.1 \\ 
\end{tblr} \end{table}


\subsection{Run-time Performance}
We conducted a runtime analysis on a laptop with an Intel Core Ultra 7 265H processor, $15$GB RAM, and an NVIDIA RTX PRO 500 Blackwell GPU with $6$GB memory. 
Table~\ref{tab:run_time} reports the inference-time comparison between NavLab and \gnetss. Although \gnetss{} has fewer trainable parameters than NavLab ($84$M versus $151$M), it requires a longer computation time of $\SI{7}{\milli\second}$ per epoch compared with $\SI{1}{\milli\second}$ for NavLab. This difference is attributed to the current implementation of Basis Net and the differentiable solver, which has not been optimised for runtime efficiency. Nevertheless, \gnetss{} remains substantially faster than the typical GNSS sampling interval of  $\SI{1}{\Hz}$, indicating that the proposed method is suitable for real-time inference.
\begin{table}[t] 
    \caption{Run-time performance} 
    \label{tab:run_time} 
    \begin{tblr}{ colspec={X[l]X[c]X[c]}, width=\textwidth, row{even}={white,font=\footnotesize}, row{odd}={bg=black!10,font=\footnotesize}, row{1}={bg=black!20,font=\bfseries\small}, hline{Z}={1pt,solid,black!60}, rowsep=1pt } Method & Parameters (M) & Runtime (ms) \\ NavLab & 151 & 1.0 $\pm$ 0.1 \\   \gnetss{} & 84 & 7.0 $\pm$ 0.3 \\ 
    \end{tblr} 
\end{table}

\section{Conclusion}
\label{sec:conclusion}

This paper presents \gnetss, a self-supervised GNSS position correction framework that eliminates the need for ground-truth receiver positions during training. 
The proposed method combines a transformer-based Basis Net with a deterministic differentiable solver, 
deriving supervision from consistency at both the measurement and trajectory levels, 
enabling scalable training from raw GNSS sequences.
Experiments on public and in-house datasets demonstrated that \gnetss{} improves positioning robustness across open-sky, urban, and forested environments. The method was particularly effective at suppressing large-error outliers, achieving substantial reductions in maximum positioning error compared with SPP and other baselines, including reductions of $38.8\%$ on GSDC and $68.2\%$ on UrbanNav in maximum 3D error relative to SPP. These results suggest that self-supervised pseudorange correction is a promising direction for robust GNSS localisation in real-world autonomous systems.
Future work will extend the framework to dual-frequency and carrier-phase measurements, with the goal of achieving robust, centimetre-level positioning accuracy in challenging real-world environments.

\nocite{*}
\printbibliography[title=References]

\end{document}